\pdfoutput=1

\documentclass[11pt]{article}

\usepackage[final]{acl}

\usepackage{times}
\usepackage{latexsym}
\usepackage{booktabs}
\usepackage{amsmath}
\usepackage{colortbl}
\usepackage{hyperref} 
\usepackage{cleveref} 
\usepackage{tcolorbox}
\usepackage[T1]{fontenc}

\usepackage[utf8]{inputenc}

\usepackage{microtype}

\usepackage{inconsolata}

\usepackage{graphicx}
\usepackage{multirow}
\title{ChartCoder: Advancing Multimodal Large Language Model for Chart-to-Code Generation}

\author{
    \textbf{Xuanle Zhao\textsuperscript{1,}\footnotemark[1]},
    \textbf{Xianzhen Luo\textsuperscript{2,}\footnotemark[1]},
    \textbf{Qi Shi\textsuperscript{1,}\footnotemark[2]},
    \textbf{Chi Chen\textsuperscript{1,}\footnotemark[2]}, 
    \\
    \textbf{Shuo Wang\textsuperscript{1}}, 
    \textbf{Zhiyuan Liu\textsuperscript{1}}, 
    \textbf{Maosong Sun\textsuperscript{1}}
    \\
\textbf{\textsuperscript{1}} Tsinghua University, Beijing, China
\\
\textbf{\textsuperscript{2}} Harbin Institute of Technology, Harbin, China 
\\
\texttt{2429527z@gmail.com, xzluo@ir.hit.edu.cn}
}

\begin{document}
\maketitle

\renewcommand{\thefootnote}{\fnsymbol{footnote}}
\footnotetext[1]{Equal contribution.}
\footnotetext[2]{Corresponding author.}
\renewcommand{\thefootnote}{\arabic{footnote}}

\begin{abstract}
Multimodal Large Language Models (MLLMs) have demonstrated remarkable capabilities in chart understanding tasks. 
However, interpreting charts with textual descriptions often leads to information loss, as it fails to fully capture the dense information embedded in charts. 
In contrast, parsing charts into code provides lossless representations that can effectively contain all critical details. Although existing open-source MLLMs have achieved success in chart understanding tasks, they still face two major challenges when applied to chart-to-code tasks: (1) Low executability and poor restoration of chart details in the generated code and (2) Lack of large-scale and diverse training data. To address these challenges, we propose \textbf{ChartCoder}, the first dedicated chart-to-code MLLM, which leverages Code LLMs as the language backbone to enhance the executability of the generated code. Furthermore, we introduce \textbf{Chart2Code-160k}, the first large-scale and diverse dataset for chart-to-code generation, and propose the \textbf{Snippet-of-Thought (SoT)} method, which transforms direct chart-to-code generation data into step-by-step generation. Experiments demonstrate that ChartCoder, with only 7B parameters, surpasses existing open-source MLLMs on chart-to-code benchmarks, achieving superior chart restoration and code excitability.
Our code is available at \url{https://github.com/thunlp/ChartCoder}.
\end{abstract}

\section{Introduction}
Recently, Multimodal Large Language Models (MLLMs) have demonstrated remarkable capabilities in addressing a wide range of visual tasks, such as captioning and question answering \cite{zhang2024multimodal, wang2024visionllm, bi2024visual, zhang2025branchlora}.
However, current models still face significant challenges in understanding and analyzing the dense visual information present in complex and informative images. As a significant form of information-intensive images, charts contain complex information such as data and structures, playing a pivotal role in effectively presenting details. The automation of chart comprehension and summarization has garnered significant attention from the research community.
\begin{figure}[t]
    \centering
    \includegraphics[width=8cm]{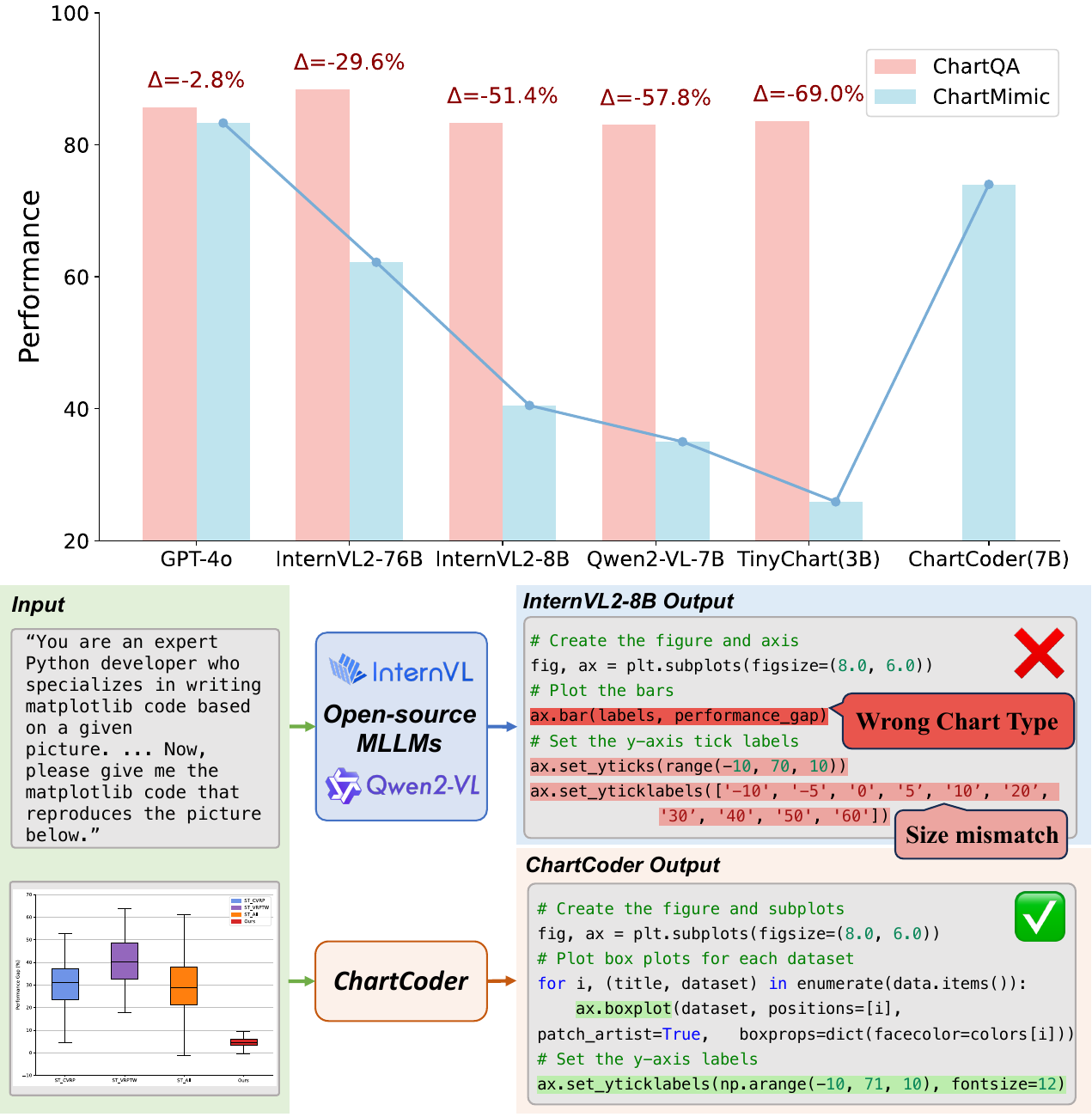}
    \caption{Comparison of existing MLLMs performance on ChartQA and ChartMimic benchmarks. In the chart-to-code task, open-source MLLMs struggle with mismatches in chart types and sizes, whereas ChartCoder generates accurate code.}
    \label{fig:overview}
    \vspace{-15pt}
\end{figure}
To advance chart understanding tasks, current studies leverage existing MLLMs and perform supervised fine-tuning (SFT) on various large-scale datasets, such as chart question answering \cite{methani2020plotqa} and chart-to-text generation \cite{kantharaj2022chart}, achieving state-of-the-art performance on existing chart understanding benchmarks. 

However, existing works generally treat charts as natural images and fine-tune models by generating natural language descriptions \cite{zhang2024tinychart,han2023chartllama,meng2024chartassisstant,bi2025prism}. This inevitably overlooks the dense information embedded within the charts, resulting in inefficient analysis and comprehension.
On the other hand, parsing a chart into code offers a lossless representation, providing a more efficient and comprehensive approach to understanding the chart by accurately capturing and summarizing all its information.
Recent works \cite{shi2024chartmimic,wu2024plot2code,xia2024chartx} have proposed various chart-to-code benchmarks, aiming to evaluate the chart reasoning abilities through code.
However, current open-source MLLMs are not well-aligned with code generation tasks \cite{zhang2024humaneval}, resulting in poor performance in parsing charts into corresponding code and limited execution rate of the generated code.
Figure \ref{fig:overview} shows that the InternVL2-8B suffers from chart type errors and coordinate size mismatches when converting boxplots to code.

To overcome the above challenges in chart-to-code generation, we first conduct an exploratory attempt by leveraging Code LLMs as the language backbone of the MLLMs and propose \textbf{ChartCoder}, the first dedicated chart-to-code MLLM, which incorporates a two-stage training paradigm that contains chart-to-text alignment and chart-to-code instruction tuning.
However, compared to chart-to-text, the available chart-to-code dataset is significantly smaller in scale, making it insufficient to support effectively supervised fine-tuning. Therefore, to address the scarcity of data for the chart-to-code domain and train our proposed ChartCoder, we propose the first large-scale, diverse and high-quality chart-to-code dataset named \textbf{Chart2Code-160k} along with the model, which contains 160k diverse chart-code pairs with 27 chart types. 
To enhance the model's capacity to capture critical information, such as chart types and data values, and strengthen its reasoning ability, we propose the \textbf{Snippet-of-Thought (SoT)} method, which emphasizes critical information and optimizes the chart-to-code reasoning process. Specifically, we sample 50k chart-code pairs from the Chart2Code-160k, then utilize Chain-of-Thought (CoT) \cite{wei2022chain} method to extend direct generation to step-by-step generation, which aims to emphasize critical information in each step. Experimental results show that by utilizing our proposed Chart2Code-160k with the SoT method, ChartCoder, which, with only 7B parameters, outperforms all open-source MLLMs across various chart-to-code benchmarks. As shown in Figure~\ref{fig:overview}, ChartCoder demonstrates a significantly higher ability to generate correct and executable code. 

In summary, the main contributions of this work are as follows:
\begin{itemize}
    \item We propose \textbf{ChartCoder}, the first chart-to-code MLLM, which leverages Code LLMs as language backbones. With only 7B parameters, ChartCoder outperforms existing open-source MLLMs on chart-to-code benchmarks.
    \item We introduce \textbf{Chart2Code-160k}, the first large-scale and diverse chart-to-code dataset, consisting of 160k chart-code pairs across 27 chart types.
    \item We propose \textbf{Snippet-of-Thought (SoT)}, transforming direct generation to step-by-step generation to emphasize critical information and enhance reasoning capabilities.  
\end{itemize}


\section{Related Works}

\subsection{Chart Understanding}
Chart understanding is a crucial area of research that encompasses both low-level and high-level tasks.
Previous approaches \cite{singh2019towards, methani2020plotqa} have typically relied on pipeline-based methods.
However, these pipeline approaches often struggle with error accumulation across different stages, which limits their overall effectiveness and flexibility.
Recent works have led to the development of end-to-end MLLMs \cite{liu2023improvedllava,liu2023llava,yu2025proglora} specifically designed for chart-related tasks. Trained on extensive chart-specific datasets, these chart-domain MLLMs \cite{xia2024chartx, zhang2024tinychart} have achieved superior performance across various chart-related tasks.
However, existing studies typically describe charts in natural language, which inevitably overlooks the dense information embedded within them, leading to inefficiencies in analysis and understanding. In contrast, code serves as a lossless representation of charts, offering a more effective and expressive approach to capturing chart information, 
thereby facilitating the solution of various chart-related tasks.

\subsection{MLLMs For Code}
Multimodal code generation has recently garnered much more attention. Several works, such as MMCode \cite{li2024mmcode} and HumanEval-V \cite{zhang2024humaneval}, have been developed to evaluate the capability of MLLMs in solving code problems that incorporate visual elements. Design2Code \cite{si2024design2code} and Web2Code \cite{yun2024web2code} evaluate the performance of MLLMs by focusing on code generation for HTML web page creation. Among the emerging tasks in this domain, chart-to-code generation has attracted significant interest as the visual elements of charts are more complex. This task challenges MLLMs to generate code that accurately reproduces a given chart or visual representation. Recent works \cite{zhang2024gpt, zhao2025chartedit} like ChartMimic \cite{shi2024chartmimic} evaluate the reasoning ability of MLLMs in this context. Similarly, Plot2Code \cite{wu2024plot2code} and ChartX \cite{xia2024chartx} also evaluate MLLMs code generation ability, especially for text and data reproducibility. To the best of our knowledge, no dedicated research has focused on solving the chart-to-code generation problem. Our work is the first to attempt to address this challenge.

\section{Chart2Code-160k Dataset}
\subsection{Direct Chart-to-code Generation}
Despite the availability of many chart-reasoning instruction-tuning datasets, there is a notable lack of datasets specifically for chart-to-code tasks. Compared to chart reasoning data, chart-to-code data have the following distinct characteristics:
(1) \textit{One-to-One Mapping}: Unlike chart reasoning datasets, which could derive multiple question-answer pairs from a single chart, chart-to-code datasets require a one-to-one correspondence, demanding a large number of chart images for training.
(2) \textit{Diversity Reflect on Charts}: 
Unlike the diversity of chart reasoning data, which can be reflected in instructions, the diversity of chart-to-code data primarily lies in the variety of chart types and structures.
(3) \textit{Syntax Constraints}: 
Unlike the flexible natural language answers in chart reasoning tasks, the output code must strictly adhere to programming syntax to ensure executability.

Therefore, collecting a large number of chart-code pairs that meet the above requirements is challenging. Recent studies have demonstrated the feasibility of generating code with LLMs \cite{xu2023wizardlm,zhang2024multimodal}. Leveraging the one-to-one mapping property of chart-to-code data, we generate code first and execute it to produce the corresponding charts. In this way, we construct the first large-scale and diverse chart-to-code dataset, named \textbf{Chart2Code-160k}. 

Specifically, we generate chart-to-code data through the following steps: First, we prompt the LLM to generate keywords within a specific domain and guide it to generate simulated data related to these keywords.
Then, to ensure the diversity of chart types, we identify 27 commonly used chart types and manually write 79 template codes for each as in-context demonstrations. These template codes contain almost all common chart formats. We further provide available functions such as \texttt{plt.text()} and parameters such as \texttt{hatch={'/'}} to encourage the generation of more diverse functions and parameters, resulting in the chart structures more diversely. To enhance the generality of generated code, LLMs are encouraged to use standard libraries such as Matplotlib and Seaborn. Additionally, we explicitly define all parameters within the code itself, eliminating the need for external files such as CSVs. This ensures that the code can be executed directly and accurately to represent the chart details. The final step involved executing the generated code to produce the corresponding chart. We utilize the above process to yield 200k code snippets for charts. After executing the code and filtering out problematic charts, such as those with excessive pixels or ticks, we construct a high-quality dataset of 160k diverse chart-to-code pairs. These pairs are formatted as multimodal instruction-tuning samples in the unified structure of \texttt{<chart, instruction, code>}.

\begin{table}[]
    \centering
    \small
    \setlength{\tabcolsep}{4pt}
    \begin{tabular}{l|ccc}
    \toprule
        Dataset &  Train/Eval & Chart Type & Number \\
    \midrule
        ChartX & Eval & 18 & 6k \\
        Plot2Code & Eval & 6 & 132 \\
        ChartMimic & Eval & 22& 2.4k \\
        ChartLlama & Train & 10 & 7.8k \\
    \midrule
        Chart2Code-160k & Train & 27 & 160k \\
    \bottomrule
    \end{tabular}
    \vspace{-5pt}
    \caption{Comparisons of existing chart-to-code datasets.}
    \label{tab:compare_chart2code}
    \vspace{-15pt}
\end{table}

\begin{figure*}[t]
    \centering
    \includegraphics[width=0.98\textwidth]{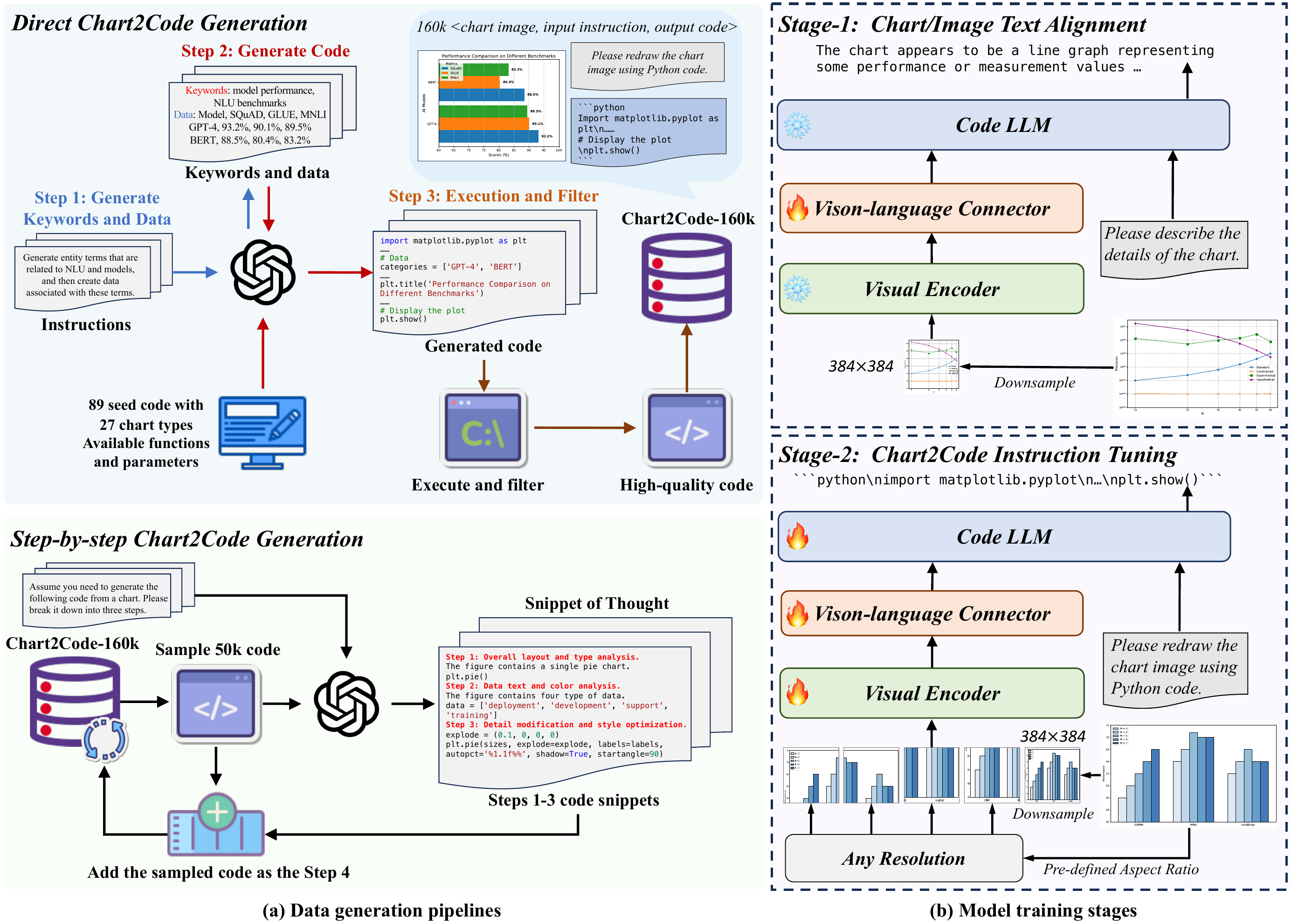}
    \vspace{-5pt}
    \caption{Illustration of Chat2Code dataset generation process and the ChartCoder training process. The dataset generation process is divided into two stages: direct generation and step-by-step generation. In the step-by-step generation, the code processed by the Snippet-of-Thought method is sampled from the Chart2Code-160k generated in the direct generation process. The training process of the ChartCoder also consists of two stages: alignment and instruction tuning.}
    \vspace{-15pt}
    \label{fig:network}
\end{figure*}

\subsection{Step-by-step Chart-to-code Generation}

Although the dataset described above includes various chart types and structures, most of the generated code follows a similar template, with only certain details (such as colors and values) providing the essential distinguishing information. This can cause chart-to-code generation models to overlook these critical details and thus produce incomplete or incorrect results. To address the above challenge and further improve the reasoning ability of MLLMs, we propose the Snippet-of-Thought (SoT) method to expand direct chart-to-code generation into step-by-step generation formats, which has demonstrated effectiveness in text-to-code generation tasks \cite{Zheng2023OutlineTD,luo2024python}.


Specifically, we adopt the SoT to imitate the human reasoning process, deriving the final code step by step. This process is divided into four steps: Step 1: Generate the chart type and layout, such as \texttt{plt.bar()} and \texttt{plt.subplot()}. Step 2: Generate the data and corresponding colors used in the chart, such as \texttt{data=[10, 15]} and \texttt{colors=['\#FF0000','\#00FF00']}. Step 3: Generate critical details of the chart, such as \texttt{hatch='/'} and \texttt{loc='upper left'}. Step 4: Generate the complete and final code.
Different from CoT and PoT, we incorporate textual explanations and code snippets for each step to emphasize key information
enhance the reasoning process and produce comprehensive outputs.

However, directly instructing the LLM to generate step-by-step code may lead to hallucinations, causing inconsistencies between intermediate code snippets and the final executable code. To maintain consistency among code snippets, we reformulated the step-by-step code data generation into a two-step process involving code generation and decomposition. We sample 50k chart-code pairs from the previously generated 160k data pairs and encourage the LLM to decompose the original code into the required textual explanations and code snippets of Steps 1–3, then concatenate the complete code in Step 4.
To further mitigate hallucinations, such as undefined values or parameters in Steps 1 and 2, we used placeholder or default parameters during code decomposition to ensure the construction of consistent and reliable step-by-step code.

\subsection{Dataset Analysis}
\textbf{Chart2Code-160k} dataset provides three key advantages: (1) \textit{The First Large-Scale Dataset}: It contains 160k data pairs for instruction tuning, significantly surpassing the size of previous datasets.
(2) \textit{Diverse Chart Structures and Types}: It includes 27 different chart types, with diverse structures enabled by a wide variety of functions and parameters in the code. (3) \textit{Syntactically Correct and Executable Code}: 
All corresponding code is syntactically correct and executable, with explicitly defined parameters that ensure precise alignment between chart structures and code representations.

The comparisons of Chart2Code-160k with relevant chart-to-code datasets are listed in \Cref{tab:compare_chart2code}. To ensure data quality, we randomly sample around 1k instances and evaluate the quality of the chart images manually during the dataset construction period. Given the strong generation capabilities of LLM, we reckon the generated charts are suitable for training purposes. Furthermore, to quantitatively evaluate the chart quality, 
we also sample 8k data pairs (5\% of the total) from Chart2Code-160k and utilize \texttt{gpt-4o-2024-08-06} to evaluate them on four criteria: \textit{Aesthetics}, \textit{Readability}, \textit{Reproducibility}, and \textit{Data Presentation Simplicity}. The results in \Cref{tab:compare_chart} show that the overall scores are broadly the same as real-world charts in ChartMimic. The detailed prompt is in the \Cref{fig:chart_quality_prompt}. Our proposed Chart2Code-160k fills the gap between chart and code, equipping the model with advanced capabilities for downstream chart tasks. 

\begin{table}[]
    \centering
    \small
    \setlength{\tabcolsep}{4pt}
    \begin{tabular}{l|c|cc}
    \toprule
        \multirow{2}{*}{Dataset} &  \multirow{2}{*}{Source} & \multicolumn{2}{c}{Chart Quality} \\
        \cmidrule{3-4}
          & & Mean $\mu$ & SD $\sigma$ \\
    \midrule
        Chart2Code-160k sample & Generated & 77.32 & 4.04 \\
        ChartMimic testmini & Real-world & 78.96 & 3.96 \\
    \bottomrule
    \end{tabular}
    \vspace{-5pt}
    \caption{Quantitative evaluation of the chart quality, comparing with real-world charts. SD is the abbreviation version for standard deviation.}
    \label{tab:compare_chart}
    \vspace{-10pt}
\end{table}


\section{ChartCoder Model}
After constructing the Chart2Code-160k, we aimed to leverage the data to enhance the capacities of MLLMs to generate code from charts. 
Unlike previous methods that rely on general LLMs with a low proportion of code in their training corpus, we pioneer the use of Code LLMs to enhance the coding abilities of MLLMs from scratch.

\subsection{Model Architecture}
Following the standard architecture of MLLMs, ChartCoder consists of three modules: a pre-trained vision encoder (SigLIP-384 \cite{Zhai2023SigmoidLF}), a vision-language connector (two-layer MLP) and a Code LLM backbone (DeepSeek Coder 6.7B \cite{guo2024deepseek}). 
The vision encoder extracts the input image into visual features, and the connector projects it into the word embedding space. LLM backbone then combines visual and textual features to generate responses.

Previous works emphasize the importance of high-resolution input for chart understanding \cite{liu2024llavanext,guo2025llava}, as details like textual words may lost in low-resolution images. However, vision transformers (ViTs) like CLIP \cite{radford2021learning} and SigLIP \cite{zhai2023sigmoid} are constrained to resolutions of $224^2$ and $384^2$ respectively, which limits their capacities to encode chart images with sufficient detail. To address this, we utilize the Any Resolution strategy \cite{liu2024llavanext} to resize and patchify chart images to ensure ChartCoder processes high-resolution chart images effectively. Specifically, the input chart image is first resized to a pre-defined optimal aspect ratio, whose height and width are integer multiples of the image resolution. The resized image is then divided into patches of standard resolution and concatenated with a directly downsampled version of the image. This approach preserves both general and detailed information without requiring the original high-resolution image to be resized into a standard square, thereby avoiding the loss of fine details. Details are shown in Figure~\ref{fig:network}.

\subsection{Model Training}
Since we propose to use Code LLMs as the language backbone to enhance the code abilities of MLLMs, existing models do not meet our requirements as their backbones are general LLMs. Thus, to align charts with text and perform supervised fine-tuning for chart-to-code tasks, we adopt the following two-stage training process.

\begin{table*}[t]
\small
\centering
\resizebox{\textwidth}{!}{
\begin{tabular}{l|r|ccc|ccc|c}
\toprule
\multirow{2}{*}{Model} & \multirow{2}{*}{Params} & \multicolumn{3}{c|}{ChartMimic} & \multicolumn{3}{c|}{Plot2Code} & ChartX \\
\cmidrule{3-9} 
  &    & Exec.Rate & Low-Level & High-Level & Pass Rate  & Text-Match  &Rating & GPT-score\\ 
\midrule
Full score & - & 100 & 100 & 100 & 100 & 100 &10 & 5 \\
\midrule
\rowcolor[gray]{0.9}
\multicolumn{9}{c}{\it{Proprietary}} \\ 
\midrule
\rowcolor[gray]{0.9} GeminiProVision & - & 68.2 & 53.8 & 53.3  & 68.2 & 53.6& 3.69 &-\\
\rowcolor[gray]{0.9} Claude-3-opus & - & 83.3 & 60.5 & 60.1 & 84.1 & 57.5 & 3.80 & -\\
\rowcolor[gray]{0.9} GPT-4V & - & 91.2 & 76.4 & 78.9 & 84.1 & 57.7 & 5.58 & 2.63\\
\rowcolor[gray]{0.9} GPT-4o & - & 93.2 & 79.0 & 83.5 & 88.6 & 56.3 & 5.71 &-\\
\midrule
\multicolumn{9}{c}{\it{Open-Source General-Domain}} \\
\midrule
DeepSeek-VL-7B & 7.3B & 41.3 & 19.0 & 17.6 & 64.4 & 32.6& 2.26 & -\\
LLaVA-Next-Mistral-7B & 7.6B & 59.7 & 20.7 & 21.3 & 72.0 & 38.7 & 2.87  &- \\
Qwen2-VL-7B & 7.0B & 67.0 & 32.9 & 35.0 & 68.2 & 33.8 & 3.10 & 1.50\\
InternVL2-4B & 4.2B & 66.2 & 33.8 & 38.4 & 66.3 & 33.4 & 2.52 & 1.57\\
InternVL2-8B & 8.1B & 61.8 & 34.4 & 38.9 & 77.3 & 37.1 & 2.78 & 1.63\\
MiniCPM-Llama3-V2.5 & 8.4B & 80.3 & 36.6 & 42.1  & 76.3 & 37.3& 2.61&1.66\\
InternVL2-26B & 26.0B & 69.3 & 41.4 & 47.4  & 81.3 & 43.1 & 3.42 & 1.70 \\
Qwen2-VL-72B & 72.0B & 73.3 &54.4  & 50.9 & 72.0 & {53.4}  & {4.26}  & 1.69  \\
InternVL2-Llama3-76B & 76.0B & {83.2} & {54.8} & {62.2}  & {85.6} & 46.6 & 3.89 & 1.74 \\
\midrule
\multicolumn{9}{c}{\it{Open-Source Chart-Domain}} \\
\midrule
ChartLlama & 13B& 57.5 & 24.8 & 28.1 & 58.4 & 40.3 & 2.32& 0.94 \\
ChartAssisstant & 13B & - & - &-  &-&-&-& 0.82 \\
TinyChart & 3B &  42.5 & 26.3 & 25.9 & 43.2 & 44.6 & 2.19 & {1.89}\\
ChartVLM-L & 14.3B & 19.5 & 15.8 & 13.9 & - & - & - &  1.58\\
\rowcolor[rgb] {1,1, 0.848} ChartCoder (Ours)  & 7.0B & \textbf{91.4} & \textbf{77.4} & \textbf{74.0}
 & \textbf{87.9} & \textbf{54.5} & \textbf{4.50} & \textbf{2.09} \\
\bottomrule 
\end{tabular}}
\vspace{-5pt}
\caption{Evaluation results of various baseline models. Unless otherwise specified, we directly use the results in the relevant benchmarks. We evaluate models that are not reported in those benchmarks. The best performances of open-source MLLMs are indicated in \textbf{bold}.}
\label{tab:main_results_direct}
\vspace{-5pt}
\end{table*}

\textbf{Chart-to-text Alignment}. 
The alignment process aims to endow the model with chart structure perception capability.
In this stage, we freeze the language and vision encoder models and pre-train the vision-language connector \cite{liu2023llava}. We collect and filter public chart corpora for alignment, which contains multiple tasks like chart caption and chart-to-table. Specifically, we use the following corpora: (1) UniChart \cite{masry2023unichart}, (2) Chart-to-Text \cite{kantharaj2022chart}, (3) SciCap \cite{hsu2021scicap}, and (4) SciCap+ \cite{yang2024scicap+}. Additionally, we incorporate the LLaVA pre-training dataset \cite{liu2023llava} and our proposed Chart2Code-160k to achieve a more balanced coverage of concepts. 

\begin{table*}[h]
\small
\centering
\begin{tabular}{l|cccccc}
\toprule
Model & Chart Types & Layout & Text Content & Data & Style & Clarity\\
\midrule
Full score & 20 &10 &20 &20 &20 & 10 \\
\midrule
\rowcolor[gray]{0.9} \rowcolor[gray]{0.9}GPT-4o& 18.96 & 9.59 & 17.16  & 15.68 & 14.66 & 8.84\\
\midrule
InternVL2-Llama3-76B & 13.06 & 8.44 & 12.59  & 10.51 & 8.74 & 7.87\\
Qwen2-VL-72B & 10.45 & 7.83 & 9.92 & 8.14 & 7.10 & 7.47  \\
InternVL2-8B & 7.20 & 6.82 & 8.81  & 5.74 & 5.42 & 6.64\\
\midrule
TinyChart & 4.16 & 5.06 & 5.22 & 2.74 & 3.21 & 5.58\\
ChartVLM-L & 0.97 & 3.53 & 2.48 & 0.81 & 0.90 &  5.25 \\
\rowcolor[rgb] {1,1, 0.848} ChartCoder (Ours) & \textbf{16.83} & \textbf{9.13} & \textbf{14.77} & \textbf{12.41} & \textbf{12.68} & \textbf{8.29} \\
\bottomrule 
\end{tabular}
\vspace{-5pt}
\caption{Detailed results of high-level scores on ChartMimic Direct Mimic task. All the subscores of ChartCoder are close to GPT-4o.}
\label{tab:main_results_high_level}
\vspace{-10pt}
\end{table*}

\textbf{Chart-to-code Instruction-tuning.}
The second stage focuses on enhancing the model's capabilities in chart-to-code tasks.
In this stage, all three modules are jointly fine-tuned with our proposed Chart2Code-160k, and additional code-related data, such as ChartQA PoT \cite{zhang2024tinychart} and ChartLlama chart-to-chart \cite{han2023chartllama}.

\section{Experiments}

\subsection{Baselines and Benchmarks}
We compare ChartCoder with existing models in three setups (1) General-domain open-source MLLMs including InternVL2(4B, 8B, 26B, 76B) \cite{chen2024internvl}, Qwen2-VL(7B, 72B) \cite{wang2024qwen2}, DeepSeek-VL-7B \cite{lu2024deepseek}, LLaVA-Next(7B) \cite{li2024llava} and MiniCPM-Llama3-V2.5 \cite{yao2024minicpm}.
(2) Proprietary models include GeminiProVision \cite{team2023gemini}, Claude-3-opus \cite{anthropic2024claude}, GPT-4V \cite{openai2023gpt4v}, and GPT-4o \cite{openai2024gpt4o}. (3) Chart-domain MLLMs including ChartLlama \cite{han2023chartllama}, ChartAssisstant \cite{meng2024chartassisstant}, Tinychart \cite{zhang2024tinychart} and ChartVLM \cite{xia2024chartx}.
All the methods are evaluated on the benchmarks ChartMimic \cite{shi2024chartmimic}, Plot2Code \cite{wu2024plot2code} and ChartX \cite{xia2024chartx}. We revise the Rating calculation in Plot2Code. The original evaluation only considers charts corresponding to executable code, which leads to higher ratings for only generating simple charts. We calculate all the results, which better reflect the impact of complex charts. For all methods, the zero-shot setting was adopted during the evaluation. Details about these benchmarks are shown in the Appendix~\ref{subsec:benchmark_details}.

\subsection{Main Results}
As indicated in \Cref{tab:main_results_direct} ChartCoder achieves the best performance among open-source MLLMs in all the chart-to-code tasks and even better than some proprietary models. Notably, on the most challenging ChartMimic task, ChartCoder surpasses leading small-scale general-domain MLLMs (<20B) such as MiniCPM-Llama3-V2.5 and InternVL2-8B with average scores of \textbf{26.7} and \textbf{34.6} respectively. The improvement achieved by ChartCoder highlights the effectiveness of our proposed Code LLM as the language backbone, combined with the Chart2Code-160k dataset, in enabling MLLMs to excel in chart understanding and code generation tasks.
In addition, ChartCoder also performs better than existing state-of-the-art large-scale MLLMs such as InternVL2-Llama3-76B and chart-domain MLLMs such as TinyChart.

We further illustrate the detailed high-level and low-level scores for the ChartMimic benchmark. The high-level score utilizes GPT-4o to evaluate the detailed similarity between the ground truth and generated chart images in six aspects: chart types, layout, text content, data, style, and clarity. The low-level score is calculated based on a comparison between the ground truth and the generated code, focusing on the code similarities in four aspects: text, layout, type, and color.

\begin{table}[]
\setlength{\tabcolsep}{4pt}
\small
\centering
\begin{tabular}{l|cccc}
\toprule
Model & Text & Layout & Type  & Color\\
\midrule
Full score & 100 & 100 & 100 & 100 \\
\midrule
\rowcolor[gray]{0.9} GPT-4o& 81.5 & 89.8 & 77.3  & 67.2\\
\midrule
InternVL2-Llama3-76B & 54.1 & 74.5 & 49.2  & 41.5\\
Qwen2-VL-72B & 43.2 & 80.5 & 54.6 & 39.4 \\
InternVL2-8B & 31.5 & 51.1 & 28.6  & 26.2 \\
\midrule
TinyChart & 9.8 & 48.2 & 32.9 & 14.2 \\
ChartVLM-L & 7.7 & 33.7 & 17.6 & 5.2 \\
\rowcolor[rgb] {1,1, 0.848} ChartCoder (Ours) & \textbf{67.2} & \textbf{95.0} & \textbf{78.5} & \textbf{69.0} \\
\bottomrule 
\end{tabular}
\vspace{-5pt}
\caption{Detailed results of low-level scores on ChartMimic Direct Mimic task. Three out of four subscores of ChartCoder are even higher than GPT-4o.}
\label{tab:main_results_low_level}
\vspace{-10pt}
\end{table}

\Cref{tab:main_results_high_level} denotes the high-level results. ChartCoder is the model most comparable to GPT-4o, as the evaluations were conducted by GPT-4o itself, suggesting the actual performance gap may not be as pronounced as it appears. Notably, ChartCoder shows the largest gap with GPT-4o in the data category, which highlights the complexity of extracting numerical values from charts, aligning with conclusions from existing chart understanding benchmarks: current MLLMs struggle to directly and accurately extract complete data from complex charts \cite{wang2024charxiv, zhang2024tinychart}.

\begin{figure*}[t]
    \centering
    \includegraphics[width=\textwidth]{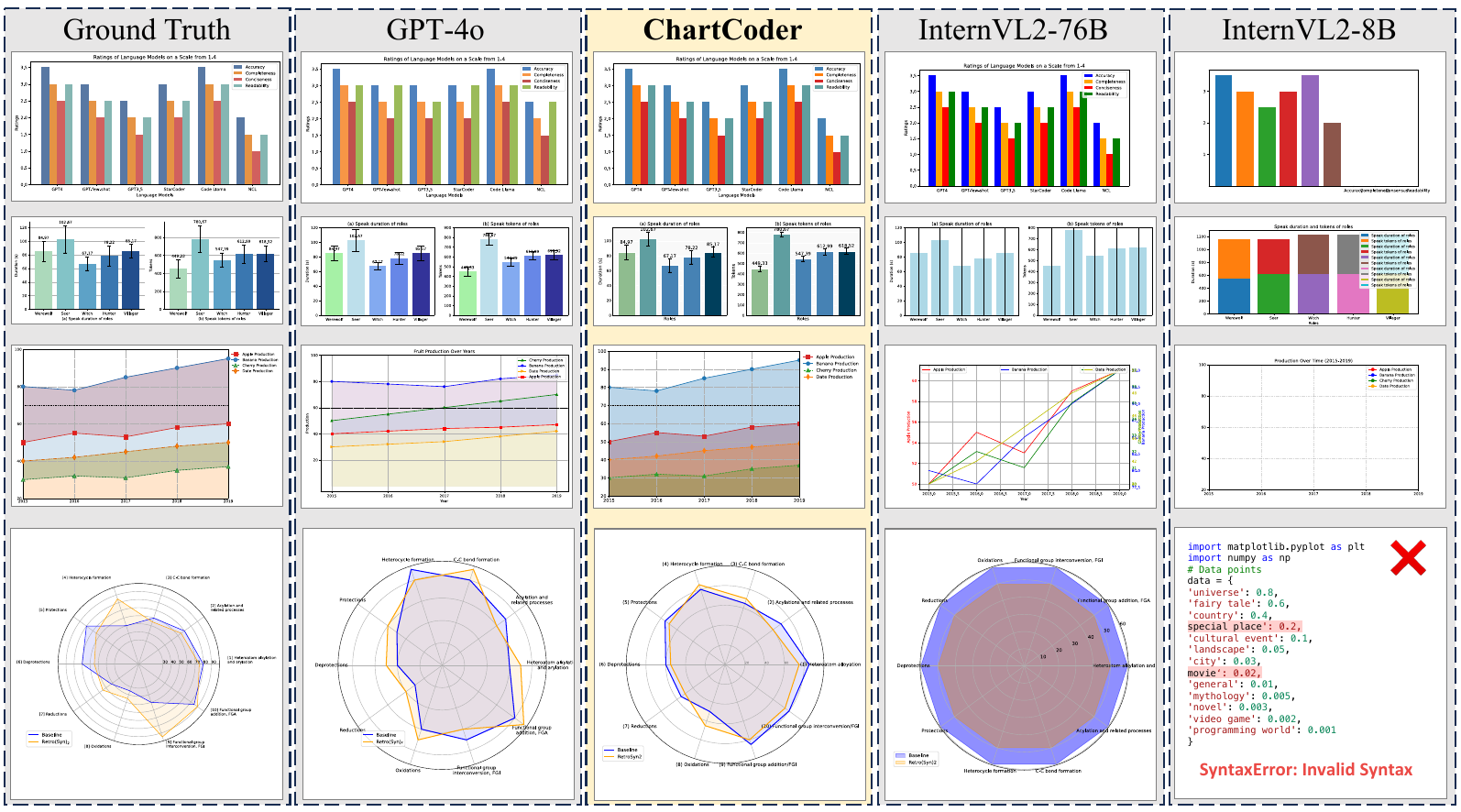}
    \caption{Generated charts of different model outputs after code execution. Our proposed ChartCoder performs significantly better than InternVL2-8B of a similar model scale.}
    \vspace{-5pt}
    \label{fig:result}
    \vspace{-5pt}
\end{figure*}

\Cref{tab:main_results_low_level} shows the low-level results. ChartCoder even slightly outperforms GPT-4o in layout, type and color, highlighting the diversity of our proposed Chart2Code-160k dataset. However, the text score of the ChartCoder is lower than GPT-4o, which is similar to the results of high-level scores. We believe this is due to the lack of specialized chart OCR-oriented training for our model. Nevertheless, our text accuracy still surpasses that of open-source models, indicating the effectiveness of our proposed ChartCoder model and Chart2Code-160k dataset. We further present some case studies on ChartMimic and compare ChartCoder with existing MLLMs. The results are shown in Figure~\ref{fig:result}, the outputs of ChartCoder are much more similar to the ground truth chart than open-source models.

\begin{table}[]
\setlength\tabcolsep{3pt}
\small
\centering
\begin{tabular}{c|ccc}
\toprule
\multirow{2.4}{*}{Methods} & \multicolumn{3}{c}{ChartMimic} \\
 & \multicolumn{1}{c|}{Exec.Rate} & \multicolumn{1}{c|}{Low-Level} & \multicolumn{1}{c}{High-Level}\\ 
\midrule
ChartCoder&  91.4 & 77.4 & 74.0 \\
\midrule
\multicolumn{4}{c}{\textit{Code LLM $\rightarrow$ General LLM} } \\
\midrule
DeepSeek LLM   & 80.6 & 61.4 & 63.4    \\
$\triangle$ & {\cellcolor[rgb] {1,0.598,0.598}}-10.8 & {\cellcolor[rgb] {1,0.402,0.402}}-16.0 & {\cellcolor[rgb] {1,0.600,0.600}}-10.6\\
\midrule
\multicolumn{4}{c}{\textit{Different Visual Encoders} } \\
\midrule
CLIP-336  & 91.6 & 77.3 &  70.3 \\
$\triangle$ & {\cellcolor[rgb]{0.954,0.954,1.0}}+0.2 & {\cellcolor[rgb]{1,0.964,0.964}}-0.1 & {\cellcolor[rgb]{1,0.872,0.872}}-3.7 \\
\midrule
\multicolumn{4}{c}{\textit{Without Step-by-step Generation}} \\
\midrule
w/o SoT   & 89.2 & 70.1 & 65.4  \\
$\triangle$  & {\cellcolor[rgb]{1,0.904,0.904}}-2.2&{\cellcolor[rgb]{1,0.742,0.742}}-7.3& {\cellcolor[rgb] {1,0.701,0.701}}-8.6\\
\midrule
\multicolumn{4}{c}{ \textit{Open-source MLLM Finetund on Chart2Code-160k}} \\
\midrule
Qwen2-VL-7B & 67.0 &32.9 &35.0 \\
\midrule
Finetuned Model & 83.6 &  73.4 & 68.2 \\
$\triangle$ & {\cellcolor[rgb]{0.824,0.824,1}}+16.7 & {\cellcolor[rgb]{0.502,0.502,1}}+40.5 & {\cellcolor[rgb] {0.682,0.682,1}}+33.2 \\
\bottomrule 
\end{tabular}
\vspace{-5pt}
\caption{The ablation studies on model architecture and data. The results show that the effectiveness
of our proposed code LLM backbone and dataset.}
\label{tab:ab_llm}
\vspace{-10pt}
\end{table}

\subsection{Ablation Study}
We perform extensive ablation experiments to validate the effectiveness of our proposed model and dataset. We divide the ablation study into three parts, and the results are shown in Table~\ref{tab:ab_llm}. (1) \textit{Code or general LLMs.}
To investigate whether employing Code LLMs as language backbone provides specific advantages in chart-to-code tasks and identify the nature of these potential benefits, we replace the Code LLM, DeepSeek Coder 6.7B, with general LLM, DeepSeek LLM 7B \cite{Bi2024DeepSeekLS}, maintaining the same two-stage training procedures. The result shows that compared with general LLM, utilizing code LLM as the language backbone could significantly improve the execution rate, as well as the low-level and high-level scores. We further analyze the types of errors in the code that failed to execute and find that utilizing code LLMs significantly reduces syntax errors like missing closing quotation marks and type errors like incorrect argument type.
(2) \textit{Resolution of vision encoders.}
Previous studies have indicated that performance on chart understanding tasks is resolution-dependent, with lower resolutions negatively impacting model performance \cite{liu2024llavanext}. To verify whether resolution affects chart-to-code tasks, we replace SigLIP-384 with CLIP-336 and maintain the other setting. The result shows that the resolution of the vision encoder generally does not affect the output code execution rate but slightly influences the high-level chart similarity. Through our analysis, we find that, similar to the challenges in chart understanding, this issue is caused by the negative impact of low resolution on the recognition of text and special symbols. However, as we utilize the Any Resolution strategy, this impact has been reduced significantly.

(3) \textit{Dataset effectiveness.} We design two scenarios to illustrate our proposed Chart2Code-160k dataset. Firstly, to evaluate our proposed SoT method to emphasize the critical information in the chart, we remove the 50k step-by-step generation data and train the model using only the direct generation data. The result shows it influences the low-level and high-level scores notably, especially in text content and data, which shows the role of emphasising critical information. Secondly, we select Qwen2-VL-7B as the baseline of open-source MLLM and directly fine-tune it on our proposed Chart2Code-160k datasets. The result illustrates that after fine-tuning, the performance improves significantly on all the metrics, demonstrating the effectiveness of Chart2Code-160k. 

\begin{table}[t]
\centering
\small
\begin{tabular}{l|ccc}
\toprule
Model & Image & Image+Code & $\triangle$ \\
\midrule
MiniCPM-Llama3-V2.5 & 0.76 & 0.81 & 6.5\%\\
InternVL2-8B & 0.79 & 0.82 & 3.8\%\\
\bottomrule 
\end{tabular}
\vspace{-5pt}
\caption{Comparison of the impact of using code as auxiliary contexts on the MMC True/False task.}
\label{tab:ab_reasoning}
\vspace{-10pt}
\end{table}

\subsection{Analysis}
We further evaluate the role of code in the chart understanding task. We use two MLLMs to evaluate two input forms, Image only and Image with Code, on the MMC True/False benchmark \cite{liu2023mmc}. The result in Table~\ref{tab:ab_reasoning} shows that using code helps the model better understand chart details, especially the chart types and the data they contain. A case study is shown in Figure~\ref{fig:reasoning}. Also, we utilize LLM to evaluate the readability of ChartCoder output code, and details are in the \Cref{app:out_code}.


\section{Conclusion}
This work aims to tackle the challenge of chart-to-code tasks with MLLMs. First, we propose the ChartCoder, which utilizes Code LLM as the language backbone dedicated to chart-to-code tasks. Second, to solve the scarcity of chart-to-code data, we present the first large-scale and diverse chart-to-code dataset, Chart2Code-160k. Finally, to emphasize the key information, we propose the Snippet-of-Thought (SoT) method to generate step-by-step data. Experiments show that ChartCoder outperforms existing open-source MLLMs. 
\section*{Limitation}
Our study is comprehensive but has certain limitations we aim to address in future research. Due to constraints in computational resources, we only trained ChartCoder with 7B parameters, which has demonstrated sufficiently good results for now. A larger model could potentially achieve even better performance. Future work may explore more complex and diverse charts and codes while experimenting with other image types like HTML to develop a comprehensive multi-modal code large language model.

\section*{Ethical Statement}
Our research employs publicly available models and datasets with proper citations. This approach minimizes the risk of generating toxic content, leveraging the widely used and non-toxic nature of our datasets and prompts.

\section*{Acknowledgment}
The work is initiated and supported by the AI9Stars Team.

\bibliography{custom}

\begin{thebibliography}{49}
\providecommand{\natexlab}[1]{#1}

\bibitem[{Anthropic(2024)}]{anthropic2024claude}
Anthropic. 2024.
\newblock \href {https://www.anthropic.com/news/claude-3-family} {Introducing the next generation of claude}.

\bibitem[{Bi et~al.(2024{\natexlab{a}})Bi, Chen, Chen, Chen, Dai, Deng, Ding, Dong, Du, Fu, Gao, Gao, Gao, Ge, Guan, Guo, Guo, Hao, Hao, He, Hu, Huang, Li, Li, Li, Li, Li, Liang, Lin, Liu, Liu), Liu, Liu, Liu, Liu, Lu, Lu, Luo, Ma, Nie, Pei, Piao, Qiu, Qu, Ren, Ren, Ruan, Sha, Shao, Song, Su, Sun, Sun, Tang, Wang, Wang, Wang, Wang, Wang, Wu, Wu, Xie, Xie, Xie, Xiong, Xu, Xu, Xu, Yang, mei You, Yu, yuan Yu, Zhang, Zhang, Zhang, Zhang, Zhang, Zhang, Zhang, Zhang, Zhao, Zhao, Zhou, Zhou, Zhu, and Zou}]{Bi2024DeepSeekLS}
DeepSeek-AI~Xiao Bi, Deli Chen, Guanting Chen, Shanhuang Chen, Damai Dai, Chengqi Deng, Honghui Ding, Kai Dong, Qiushi Du, Zhe Fu, Huazuo Gao, Kaige Gao, Wenjun Gao, Ruiqi Ge, Kang Guan, Daya Guo, Jianzhong Guo, Guangbo Hao, Zhewen Hao, Ying He, Wen-Hui Hu, Panpan Huang, Erhang Li, Guowei Li, Jiashi Li, Yao Li, Y.~K. Li, Wenfeng Liang, Fangyun Lin, Aixin Liu, Bo~Liu~(Benjamin Liu), Wen Liu, Xiaodong Liu, Xin Liu, Yiyuan Liu, Haoyu Lu, Shanghao Lu, Fuli Luo, Shirong Ma, Xiaotao Nie, Tian Pei, Yishi Piao, Junjie Qiu, Hui Qu, Tongzheng Ren, Zehui Ren, Chong Ruan, Zhangli Sha, Zhihong Shao, Jun-Mei Song, Xuecheng Su, Jingxiang Sun, Yaofeng Sun, Min Tang, Bing-Li Wang, Peiyi Wang, Shiyu Wang, Yaohui Wang, Yongji Wang, Tong Wu, Yu~Wu, Xin Xie, Zhenda Xie, Ziwei Xie, Yi~Xiong, Hanwei Xu, Ronald~X Xu, Yanhong Xu, Dejian Yang, Yu~mei You, Shuiping Yu, Xin yuan Yu, Bo~Zhang, Haowei Zhang, Lecong Zhang, Liyue Zhang, Mingchuan Zhang, Minghu Zhang, Wentao Zhang, Yichao Zhang, Chenggang Zhao, Yao Zhao, Shangyan Zhou,
  Shunfeng Zhou, Qihao Zhu, and Yuheng Zou. 2024{\natexlab{a}}.
\newblock Deepseek llm: Scaling open-source language models with longtermism.
\newblock \emph{ArXiv}, abs/2401.02954.

\bibitem[{Bi et~al.(2025)Bi, Wang, Yan, Xiao, Hecker, Tresp, and Ma}]{bi2025prism}
Jinhe Bi, Yifan Wang, Danqi Yan, Xun Xiao, Artur Hecker, Volker Tresp, and Yunpu Ma. 2025.
\newblock Prism: Self-pruning intrinsic selection method for training-free multimodal data selection.
\newblock \emph{arXiv preprint arXiv:2502.12119}.

\bibitem[{Bi et~al.(2024{\natexlab{b}})Bi, Wang, Chen, Xiao, Hecker, Tresp, and Ma}]{bi2024visual}
Jinhe Bi, Yujun Wang, Haokun Chen, Xun Xiao, Artur Hecker, Volker Tresp, and Yunpu Ma. 2024{\natexlab{b}}.
\newblock Visual instruction tuning with 500x fewer parameters through modality linear representation-steering.
\newblock \emph{arXiv preprint arXiv:2412.12359}.

\bibitem[{Chen et~al.(2024)Chen, Wu, Wang, Su, Chen, Xing, Zhong, Zhang, Zhu, Lu et~al.}]{chen2024internvl}
Zhe Chen, Jiannan Wu, Wenhai Wang, Weijie Su, Guo Chen, Sen Xing, Muyan Zhong, Qinglong Zhang, Xizhou Zhu, Lewei Lu, et~al. 2024.
\newblock Internvl: Scaling up vision foundation models and aligning for generic visual-linguistic tasks.
\newblock In \emph{Proceedings of the IEEE/CVF Conference on Computer Vision and Pattern Recognition}, pages 24185--24198.

\bibitem[{Guo et~al.(2024)Guo, Zhu, Yang, Xie, Dong, Zhang, Chen, Bi, Wu, Li et~al.}]{guo2024deepseek}
Daya Guo, Qihao Zhu, Dejian Yang, Zhenda Xie, Kai Dong, Wentao Zhang, Guanting Chen, Xiao Bi, Yu~Wu, YK~Li, et~al. 2024.
\newblock Deepseek-coder: When the large language model meets programming--the rise of code intelligence.
\newblock \emph{arXiv preprint arXiv:2401.14196}.

\bibitem[{Guo et~al.(2025)Guo, Xu, Yao, Cui, Ni, Ge, Chua, Liu, and Huang}]{guo2025llava}
Zonghao Guo, Ruyi Xu, Yuan Yao, Junbo Cui, Zanlin Ni, Chunjiang Ge, Tat-Seng Chua, Zhiyuan Liu, and Gao Huang. 2025.
\newblock Llava-uhd: an lmm perceiving any aspect ratio and high-resolution images.
\newblock In \emph{European Conference on Computer Vision}, pages 390--406. Springer.

\bibitem[{Han et~al.(2023)Han, Zhang, Chen, Yang, Wang, Yu, Fu, and Zhang}]{han2023chartllama}
Yucheng Han, Chi Zhang, Xin Chen, Xu~Yang, Zhibin Wang, Gang Yu, Bin Fu, and Hanwang Zhang. 2023.
\newblock Chartllama: A multimodal llm for chart understanding and generation.
\newblock \emph{arXiv preprint arXiv:2311.16483}.

\bibitem[{Hsu et~al.(2021)Hsu, Giles, and Huang}]{hsu2021scicap}
Ting-Yao Hsu, C~Lee Giles, and Ting-Hao'Kenneth' Huang. 2021.
\newblock Scicap: Generating captions for scientific figures.
\newblock \emph{arXiv preprint arXiv:2110.11624}.

\bibitem[{Kantharaj et~al.(2022)Kantharaj, Leong, Lin, Masry, Thakkar, Hoque, and Joty}]{kantharaj2022chart}
Shankar Kantharaj, Rixie Tiffany~Ko Leong, Xiang Lin, Ahmed Masry, Megh Thakkar, Enamul Hoque, and Shafiq Joty. 2022.
\newblock Chart-to-text: A large-scale benchmark for chart summarization.
\newblock \emph{arXiv preprint arXiv:2203.06486}.

\bibitem[{Li et~al.(2024{\natexlab{a}})Li, Zhang, Zhang, Zhang, Li, Li, Ma, and Li}]{li2024llava}
Feng Li, Renrui Zhang, Hao Zhang, Yuanhan Zhang, Bo~Li, Wei Li, Zejun Ma, and Chunyuan Li. 2024{\natexlab{a}}.
\newblock Llava-next-interleave: Tackling multi-image, video, and 3d in large multimodal models.
\newblock \emph{arXiv preprint arXiv:2407.07895}.

\bibitem[{Li et~al.(2024{\natexlab{b}})Li, Tian, Hu, Luo, Huang, and Ma}]{li2024mmcode}
Kaixin Li, Yuchen Tian, Qisheng Hu, Ziyang Luo, Zhiyong Huang, and Jing Ma. 2024{\natexlab{b}}.
\newblock Mmcode: Benchmarking multimodal large language models for code generation with visually rich programming problems.
\newblock In \emph{Findings of the Association for Computational Linguistics: EMNLP 2024}, pages 736--783.

\bibitem[{Liu et~al.(2023{\natexlab{a}})Liu, Wang, Yao, Chen, Song, Cho, Yacoob, and Yu}]{liu2023mmc}
Fuxiao Liu, Xiaoyang Wang, Wenlin Yao, Jianshu Chen, Kaiqiang Song, Sangwoo Cho, Yaser Yacoob, and Dong Yu. 2023{\natexlab{a}}.
\newblock Mmc: Advancing multimodal chart understanding with large-scale instruction tuning.
\newblock \emph{arXiv preprint arXiv:2311.10774}.

\bibitem[{Liu et~al.(2023{\natexlab{b}})Liu, Li, Li, and Lee}]{liu2023improvedllava}
Haotian Liu, Chunyuan Li, Yuheng Li, and Yong~Jae Lee. 2023{\natexlab{b}}.
\newblock Improved baselines with visual instruction tuning.

\bibitem[{Liu et~al.(2024)Liu, Li, Li, Li, Zhang, Shen, and Lee}]{liu2024llavanext}
Haotian Liu, Chunyuan Li, Yuheng Li, Bo~Li, Yuanhan Zhang, Sheng Shen, and Yong~Jae Lee. 2024.
\newblock \href {https://llava-vl.github.io/blog/2024-01-30-llava-next/} {Llava-next: Improved reasoning, ocr, and world knowledge}.

\bibitem[{Liu et~al.(2023{\natexlab{c}})Liu, Li, Wu, and Lee}]{liu2023llava}
Haotian Liu, Chunyuan Li, Qingyang Wu, and Yong~Jae Lee. 2023{\natexlab{c}}.
\newblock Visual instruction tuning.

\bibitem[{Lu et~al.(2024)Lu, Liu, Zhang, Wang, Dong, Liu, Sun, Ren, Li, Yang et~al.}]{lu2024deepseek}
Haoyu Lu, Wen Liu, Bo~Zhang, Bingxuan Wang, Kai Dong, Bo~Liu, Jingxiang Sun, Tongzheng Ren, Zhuoshu Li, Hao Yang, et~al. 2024.
\newblock Deepseek-vl: towards real-world vision-language understanding.
\newblock \emph{arXiv preprint arXiv:2403.05525}.

\bibitem[{Luo et~al.(2024)Luo, Zhu, Zhang, Qin, Zhang, Yang, Xu, and Che}]{luo2024python}
Xianzhen Luo, Qingfu Zhu, Zhiming Zhang, Libo Qin, Xuanyu Zhang, Qing Yang, Dongliang Xu, and Wanxiang Che. 2024.
\newblock Python is not always the best choice: Embracing multilingual program of thoughts.
\newblock In \emph{Proceedings of the 2024 Conference on Empirical Methods in Natural Language Processing}, pages 7185--7212.

\bibitem[{Masry et~al.(2023)Masry, Kavehzadeh, Do, Hoque, and Joty}]{masry2023unichart}
Ahmed Masry, Parsa Kavehzadeh, Xuan~Long Do, Enamul Hoque, and Shafiq Joty. 2023.
\newblock Unichart: A universal vision-language pretrained model for chart comprehension and reasoning.
\newblock \emph{arXiv preprint arXiv:2305.14761}.

\bibitem[{Meng et~al.(2024)Meng, Shao, Lu, Gao, Zhang, Qiao, and Luo}]{meng2024chartassisstant}
Fanqing Meng, Wenqi Shao, Quanfeng Lu, Peng Gao, Kaipeng Zhang, Yu~Qiao, and Ping Luo. 2024.
\newblock Chartassisstant: A universal chart multimodal language model via chart-to-table pre-training and multitask instruction tuning.
\newblock \emph{arXiv preprint arXiv:2401.02384}.

\bibitem[{Methani et~al.(2020)Methani, Ganguly, Khapra, and Kumar}]{methani2020plotqa}
Nitesh Methani, Pritha Ganguly, Mitesh~M Khapra, and Pratyush Kumar. 2020.
\newblock Plotqa: Reasoning over scientific plots.
\newblock In \emph{Proceedings of the IEEE/CVF Winter Conference on Applications of Computer Vision}, pages 1527--1536.

\bibitem[{OpenAI(2023)}]{openai2023gpt4v}
OpenAI. 2023.
\newblock \href {https://openai.com/index/gpt-4v-system-card/} {Gpt-4v(ision) system card}.

\bibitem[{OpenAI(2024)}]{openai2024gpt4o}
OpenAI. 2024.
\newblock \href {https://openai.com/index/hello-gpt-4o} {Gpt-4o}.
\newblock Accessed: 2024-05-13.

\bibitem[{{Qwen Team}(2024)}]{qwen2.5}
{Qwen Team}. 2024.
\newblock \href {https://qwenlm.github.io/blog/qwen2.5/} {Qwen2.5: A party of foundation models}.

\bibitem[{Radford et~al.(2021)Radford, Kim, Hallacy, Ramesh, Goh, Agarwal, Sastry, Askell, Mishkin, Clark et~al.}]{radford2021learning}
Alec Radford, Jong~Wook Kim, Chris Hallacy, Aditya Ramesh, Gabriel Goh, Sandhini Agarwal, Girish Sastry, Amanda Askell, Pamela Mishkin, Jack Clark, et~al. 2021.
\newblock Learning transferable visual models from natural language supervision.
\newblock In \emph{International conference on machine learning}, pages 8748--8763. PMLR.

\bibitem[{Shi et~al.(2024)Shi, Yang, Liu, Shui, Wang, Jing, Xu, Zhu, Li, Zhang et~al.}]{shi2024chartmimic}
Chufan Shi, Cheng Yang, Yaxin Liu, Bo~Shui, Junjie Wang, Mohan Jing, Linran Xu, Xinyu Zhu, Siheng Li, Yuxiang Zhang, et~al. 2024.
\newblock Chartmimic: Evaluating lmm's cross-modal reasoning capability via chart-to-code generation.
\newblock \emph{arXiv preprint arXiv:2406.09961}.

\bibitem[{Si et~al.(2024)Si, Zhang, Yang, Liu, and Yang}]{si2024design2code}
Chenglei Si, Yanzhe Zhang, Zhengyuan Yang, Ruibo Liu, and Diyi Yang. 2024.
\newblock Design2code: How far are we from automating front-end engineering?
\newblock \emph{arXiv preprint arXiv:2403.03163}.

\bibitem[{Singh et~al.(2019)Singh, Natarajan, Shah, Jiang, Chen, Batra, Parikh, and Rohrbach}]{singh2019towards}
Amanpreet Singh, Vivek Natarajan, Meet Shah, Yu~Jiang, Xinlei Chen, Dhruv Batra, Devi Parikh, and Marcus Rohrbach. 2019.
\newblock Towards vqa models that can read.
\newblock In \emph{Proceedings of the IEEE/CVF conference on computer vision and pattern recognition}, pages 8317--8326.

\bibitem[{Team et~al.(2023)Team, Anil, Borgeaud, Alayrac, Yu, Soricut, Schalkwyk, Dai, Hauth, Millican et~al.}]{team2023gemini}
Gemini Team, Rohan Anil, Sebastian Borgeaud, Jean-Baptiste Alayrac, Jiahui Yu, Radu Soricut, Johan Schalkwyk, Andrew~M Dai, Anja Hauth, Katie Millican, et~al. 2023.
\newblock Gemini: a family of highly capable multimodal models.
\newblock \emph{arXiv preprint arXiv:2312.11805}.

\bibitem[{Wang et~al.(2024{\natexlab{a}})Wang, Bai, Tan, Wang, Fan, Bai, Chen, Liu, Wang, Ge et~al.}]{wang2024qwen2}
Peng Wang, Shuai Bai, Sinan Tan, Shijie Wang, Zhihao Fan, Jinze Bai, Keqin Chen, Xuejing Liu, Jialin Wang, Wenbin Ge, et~al. 2024{\natexlab{a}}.
\newblock Qwen2-vl: Enhancing vision-language model's perception of the world at any resolution.
\newblock \emph{arXiv preprint arXiv:2409.12191}.

\bibitem[{Wang et~al.(2024{\natexlab{b}})Wang, Chen, Chen, Wu, Zhu, Zeng, Luo, Lu, Zhou, Qiao et~al.}]{wang2024visionllm}
Wenhai Wang, Zhe Chen, Xiaokang Chen, Jiannan Wu, Xizhou Zhu, Gang Zeng, Ping Luo, Tong Lu, Jie Zhou, Yu~Qiao, et~al. 2024{\natexlab{b}}.
\newblock Visionllm: Large language model is also an open-ended decoder for vision-centric tasks.
\newblock \emph{Advances in Neural Information Processing Systems}, 36.

\bibitem[{Wang et~al.(2024{\natexlab{c}})Wang, Xia, He, Chen, Liu, Zhu, Liang, Wu, Liu, Malladi et~al.}]{wang2024charxiv}
Zirui Wang, Mengzhou Xia, Luxi He, Howard Chen, Yitao Liu, Richard Zhu, Kaiqu Liang, Xindi Wu, Haotian Liu, Sadhika Malladi, et~al. 2024{\natexlab{c}}.
\newblock Charxiv: Charting gaps in realistic chart understanding in multimodal llms.
\newblock \emph{arXiv preprint arXiv:2406.18521}.

\bibitem[{Wei et~al.(2022)Wei, Wang, Schuurmans, Bosma, Xia, Chi, Le, Zhou et~al.}]{wei2022chain}
Jason Wei, Xuezhi Wang, Dale Schuurmans, Maarten Bosma, Fei Xia, Ed~Chi, Quoc~V Le, Denny Zhou, et~al. 2022.
\newblock Chain-of-thought prompting elicits reasoning in large language models.
\newblock \emph{Advances in neural information processing systems}, 35:24824--24837.

\bibitem[{Wu et~al.(2024)Wu, Ge, Guo, Wang, Liang, Lu, Shan, and Luo}]{wu2024plot2code}
Chengyue Wu, Yixiao Ge, Qiushan Guo, Jiahao Wang, Zhixuan Liang, Zeyu Lu, Ying Shan, and Ping Luo. 2024.
\newblock Plot2code: A comprehensive benchmark for evaluating multi-modal large language models in code generation from scientific plots.
\newblock \emph{arXiv preprint arXiv:2405.07990}.

\bibitem[{Xia et~al.(2024)Xia, Zhang, Ye, Yan, Liu, Zhou, Chen, Dou, Shi, Yan et~al.}]{xia2024chartx}
Renqiu Xia, Bo~Zhang, Hancheng Ye, Xiangchao Yan, Qi~Liu, Hongbin Zhou, Zijun Chen, Min Dou, Botian Shi, Junchi Yan, et~al. 2024.
\newblock Chartx \& chartvlm: A versatile benchmark and foundation model for complicated chart reasoning.
\newblock \emph{arXiv preprint arXiv:2402.12185}.

\bibitem[{Xu et~al.(2023)Xu, Sun, Zheng, Geng, Zhao, Feng, Tao, and Jiang}]{xu2023wizardlm}
Can Xu, Qingfeng Sun, Kai Zheng, Xiubo Geng, Pu~Zhao, Jiazhan Feng, Chongyang Tao, and Daxin Jiang. 2023.
\newblock Wizardlm: Empowering large language models to follow complex instructions.
\newblock \emph{arXiv preprint arXiv:2304.12244}.

\bibitem[{Yang et~al.(2024)Yang, Dabre, Tanaka, and Okazaki}]{yang2024scicap+}
Zhishen Yang, Raj Dabre, Hideki Tanaka, and Naoaki Okazaki. 2024.
\newblock Scicap+: A knowledge augmented dataset to study the challenges of scientific figure captioning.
\newblock \emph{Journal of Natural Language Processing}, 31(3):1140--1165.

\bibitem[{Yao et~al.(2024)Yao, Yu, Zhang, Wang, Cui, Zhu, Cai, Li, Zhao, He et~al.}]{yao2024minicpm}
Yuan Yao, Tianyu Yu, Ao~Zhang, Chongyi Wang, Junbo Cui, Hongji Zhu, Tianchi Cai, Haoyu Li, Weilin Zhao, Zhihui He, et~al. 2024.
\newblock Minicpm-v: A gpt-4v level mllm on your phone.
\newblock \emph{arXiv preprint arXiv:2408.01800}.

\bibitem[{Yu et~al.(2025)Yu, Zhang, Ren, Zhao, Chen, and Chu}]{yu2025proglora}
Yahan Yu, Duzhen Zhang, Yong Ren, Xuanle Zhao, Xiuyi Chen, and Chenhui Chu. 2025.
\newblock Progressive lora for multimodal continual instruction tuning.
\newblock In \emph{Findings of the Association for Computational Linguistics ACL 2025}.

\bibitem[{Yun et~al.(2024)Yun, Lin, Thushara, Bhat, Wang, Jiang, Deng, Wang, Tao, Li et~al.}]{yun2024web2code}
Sukmin Yun, Haokun Lin, Rusiru Thushara, Mohammad~Qazim Bhat, Yongxin Wang, Zutao Jiang, Mingkai Deng, Jinhong Wang, Tianhua Tao, Junbo Li, et~al. 2024.
\newblock Web2code: A large-scale webpage-to-code dataset and evaluation framework for multimodal llms.
\newblock \emph{arXiv preprint arXiv:2406.20098}.

\bibitem[{Zhai et~al.(2023{\natexlab{a}})Zhai, Mustafa, Kolesnikov, and Beyer}]{Zhai2023SigmoidLF}
Xiaohua Zhai, Basil Mustafa, Alexander Kolesnikov, and Lucas Beyer. 2023{\natexlab{a}}.
\newblock Sigmoid loss for language image pre-training.
\newblock \emph{2023 IEEE/CVF International Conference on Computer Vision (ICCV)}, pages 11941--11952.

\bibitem[{Zhai et~al.(2023{\natexlab{b}})Zhai, Mustafa, Kolesnikov, and Beyer}]{zhai2023sigmoid}
Xiaohua Zhai, Basil Mustafa, Alexander Kolesnikov, and Lucas Beyer. 2023{\natexlab{b}}.
\newblock Sigmoid loss for language image pre-training.
\newblock In \emph{Proceedings of the IEEE/CVF International Conference on Computer Vision}, pages 11975--11986.

\bibitem[{Zhang et~al.(2025)Zhang, Ren, Li, Yu, Dong, Li, Ji, and Bai}]{zhang2025branchlora}
Duzhen Zhang, Yong Ren, Zhong-Zhi Li, Yahan Yu, Jiahua Dong, Chenxing Li, Zhilong Ji, and Jinfeng Bai. 2025.
\newblock Enhancing multimodal continual instruction tuning with branchlora.
\newblock In \emph{Proceedings of the 63rd Annual Meeting of the Association for Computational Linguistics (Volume 1: Long Papers)}.

\bibitem[{Zhang et~al.(2024{\natexlab{a}})Zhang, Wu, Bai, Lin, Li, Yu, Wang, Chen, and Keung}]{zhang2024humaneval}
Fengji Zhang, Linquan Wu, Huiyu Bai, Guancheng Lin, Xiao Li, Xiao Yu, Yue Wang, Bei Chen, and Jacky Keung. 2024{\natexlab{a}}.
\newblock Humaneval-v: Evaluating visual understanding and reasoning abilities of large multimodal models through coding tasks.
\newblock \emph{arXiv preprint arXiv:2410.12381}.

\bibitem[{Zhang et~al.(2024{\natexlab{b}})Zhang, Hu, Xu, Yan, Xu, Jin, Zhang, and Huang}]{zhang2024tinychart}
Liang Zhang, Anwen Hu, Haiyang Xu, Ming Yan, Yichen Xu, Qin Jin, Ji~Zhang, and Fei Huang. 2024{\natexlab{b}}.
\newblock Tinychart: Efficient chart understanding with visual token merging and program-of-thoughts learning.
\newblock \emph{arXiv preprint arXiv:2404.16635}.

\bibitem[{Zhang et~al.(2024{\natexlab{c}})Zhang, Cheng, He, Wang, Shen, Tan, Hou, He, Ma, Lu et~al.}]{zhang2024multimodal}
Wenqi Zhang, Zhenglin Cheng, Yuanyu He, Mengna Wang, Yongliang Shen, Zeqi Tan, Guiyang Hou, Mingqian He, Yanna Ma, Weiming Lu, et~al. 2024{\natexlab{c}}.
\newblock Multimodal self-instruct: Synthetic abstract image and visual reasoning instruction using language model.
\newblock \emph{arXiv preprint arXiv:2407.07053}.

\bibitem[{Zhang et~al.(2024{\natexlab{d}})Zhang, Ma, and Vosoughi}]{zhang2024gpt}
Zhehao Zhang, Weicheng Ma, and Soroush Vosoughi. 2024{\natexlab{d}}.
\newblock Is gpt-4v (ision) all you need for automating academic data visualization? exploring vision-language models’ capability in reproducing academic charts.
\newblock In \emph{Findings of the Association for Computational Linguistics: EMNLP 2024}, pages 8271--8288.

\bibitem[{Zhao et~al.(2025)Zhao, Liu, Yang, Luo, Zeng, Li, Shi, and Chen}]{zhao2025chartedit}
Xuanle Zhao, Xuexin Liu, Haoyue Yang, Xianzhen Luo, Fanhu Zeng, Jianling Li, Qi~Shi, and Chi Chen. 2025.
\newblock Chartedit: How far are mllms from automating chart analysis? evaluating mllms' capability via chart editing.
\newblock \emph{arXiv preprint arXiv:2505.11935}.

\bibitem[{Zheng et~al.(2023)Zheng, Sharan, Jaiswal, Wang, Xi, Xu, and Wang}]{Zheng2023OutlineTD}
Wenqing Zheng, S~P Sharan, Ajay Jaiswal, Kevin Wang, Yihan Xi, Dejia Xu, and Zhangyang Wang. 2023.
\newblock Outline, then details: Syntactically guided coarse-to-fine code generation.
\newblock In \emph{International Conference on Machine Learning}.

\end{thebibliography}

\clearpage
\appendix

\section{Appendix}
\label{sec:appendix}
\subsection{Implementation Details}
In the data generation stage, we utilize \texttt{gpt-4o-2024-08-06} as the LLM for both direct and step-by-step generation processes.

In the training stage, ChartCoder is initialized with SigLIP-384 \cite{radford2021learning} as the vision encoder and DeepSeek Coder 6.7B \cite{guo2024deepseek} as the large language model. The whole training process is divided into alignment and instruction tuning. During the alignment stage, we only train the vision-language connector with the chart-to-text alignment data. The learning rate is set to 1e-3.
In the instruction tuning stage, we train the entire model for 1 epoch with a batchsize of 128.  The learning rate of SigLIP and other modules are 5e-6 and 1e-5 respectively, with a warmup at the beginning of 3\%, then decays to 0 at the end of training.  The alignment and instruction tuning processes cost 12 and 5 hours on 32 Tesla A100 GPUs with 80 GB VRAMs.

\subsection{Benchmark Details}
\label{subsec:benchmark_details}
\textbf{ChartMimic} \cite{shi2024chartmimic} focuses on evaluating the ability of MLLMs to redraw charts from ArXiv papers, emphasizing the preservation of the original style and appearance. It consists of two subsets: testmini and test. Following the settings in the original paper, we adopt the Direct Mimic task on the testmini subset as the default evaluation standard, reporting execution success rates alongside low-level and high-level scores.

\textbf{Plot2Code} \cite{wu2024plot2code} aims to evaluate models' abilities to generate code corresponding to charts from the available Matplotlib galleries, with a focus on textual similarity. We evaluate models on its Direct Asking task using three metrics: Pass Rate, Text-Match, and Rating.

\textbf{ChartX} \cite{xia2024chartx} contains various tasks with synthesis chart images, including Question Answering, Summarization, Description and Redrawing. We choose the Redrawing task and report the GPT score as the metrics in ChartX.

\subsection{More Ablation Studies}
We also perform more ablation studies on the language backbone and further choose Qwen2.5-7B and Qwen2.5 Coder-7B \cite{qwen2.5}  for comparison. The results also show that using Code LLM as the language backbone is better than using general LLM. However, we find that using the Qwen2.5 Coder as the backbone does not perform as well as using the DeepSeek Coder. 
This observation seems counterintuitive, as the official evaluation suggests that the performance of the Qwen2.5 Coder is better than the DeepSeek Coder. We analyze experimental results and find that the code generated by Qwen2.5 is more standardized. For instance, the DeepSeek Coder backbone tends to use \texttt{ax[0], ax[1]}, while the Qwen2.5 Coder backbone prefers a more standardized approach, such as using \texttt{for i in range(2): ax[i]}. However, in some complex scenarios, using a \texttt{for} loop may lead to errors, such as \texttt{ax[0]} and \texttt{ax[1]} do not have same number of bars.

\begin{table}[]
\setlength{\tabcolsep}{3pt}
\small
\centering
\begin{tabular}{c|ccc}
\toprule
\multirow{2.4}{*}{Methods} & \multicolumn{3}{c}{ChartMimic} \\
\cmidrule{2-4} 
 & \multicolumn{1}{c|}{Exec.Rate} & \multicolumn{1}{c|}{Low-Level} & \multicolumn{1}{c}{High-Level}\\ 
\midrule
ChartCoder & 91.4 & 77.4 & 74.0 \\
\midrule
\multicolumn{4}{c}{\textit{Replace Language Backbone} } \\
\midrule
Qwen2.5   & 88.1 & 73.4 & 67.9    \\
$\triangle$ & {\cellcolor[rgb]{1,0.894,0.894}}-3.3 & {\cellcolor[rgb]{1,0.914,0.914}}-4.0 & {\cellcolor[rgb] {1,0.502,0.502}}-6.1 \\
Qwen2.5 Coder   & 90.3 & 76.8 &  69.7 \\
$\triangle$& {\cellcolor[rgb]{1,0.894,0.894}}-1.1 & {\cellcolor[rgb]{1,0.894,0.894}}-0.6 &  {\cellcolor[rgb]{1,0.592,0.592}}-4.3\\
\bottomrule 
\end{tabular}
\caption{The ablation studies on model architecture and data. The results show that the effectiveness
of our proposed model architecture and dataset.}
\label{tab:ab_qwen}
\end{table}

\subsection{Output Code Analysis}\label{app:out_code}
\begin{figure}[t]
    \centering
    \includegraphics[width=7.5cm]{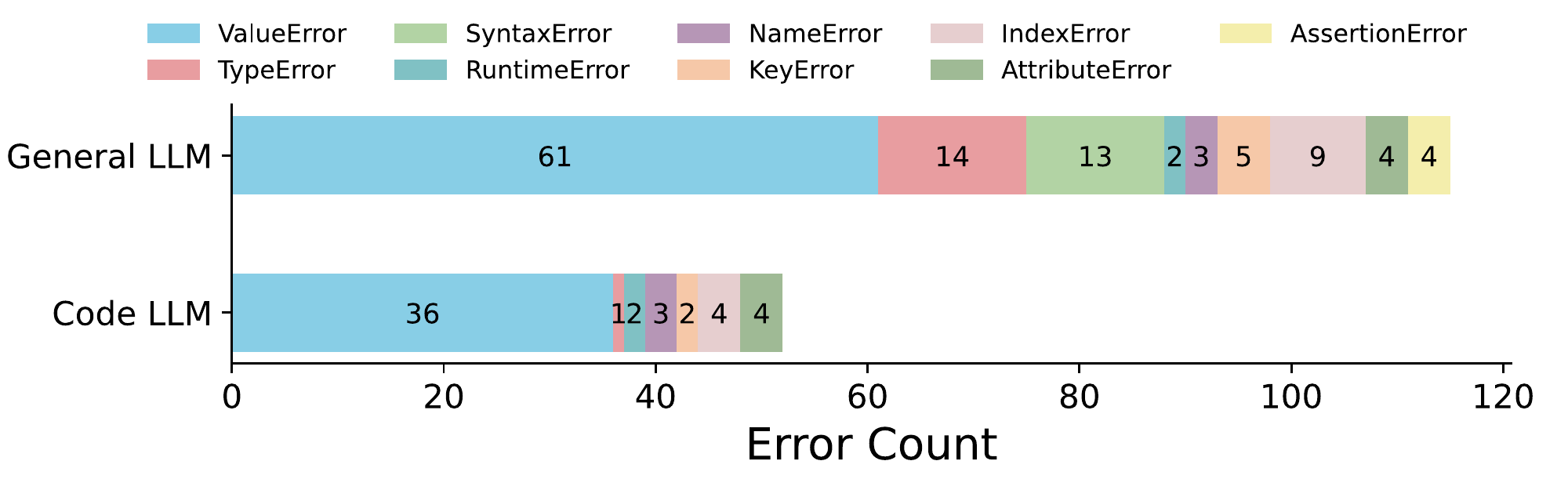}
    \caption{Comparison of error types on ChartMimic direct generation tasks with code and general LLMs as language backbone, respectively.}
    \vspace{-5pt}
    \label{fig:ablation}
\end{figure}

To evaluate the output code readability, we conduct an ablation experiment, utilizing \texttt{gpt-4o-2024-08-06} to evaluate the output code readability. We evaluate four aspects of the generated code, including \textit{Naming Conventions}, \textit{Code Structure}, \textit{Comments}, and \textit{Logical Clarity}, with a total score of 100. We choose the generated code from the ChartMimic task (ChartCoder output) and the ground truth code (human-annotated) in the ChartMimic dataset. The results are as shown in \Cref{tab:code_comparison}. We also evaluate the error types on ChartMimic direct generation tasks with code and general LLMs as the language backbone. The results are shown in \Cref{fig:ablation}.

\begin{table}[]
    \centering
    \small
    \setlength{\tabcolsep}{3pt}
    \begin{tabular}{l|c|cc}
    \toprule
    \multirow{2}{*}{Dataset} &  \multirow{2}{*}{Source} & \multicolumn{2}{c}{Code Readability} \\
    \cmidrule{3-4}
      & & Mean $\mu$ & SD $\sigma$ \\
    \hline
    Qwen2-VL-7B Output & Model generated & 82.48 & 6.81 \\
    ChartCoder Output & Model generated & 85.22 & 6.78 \\
    ChartMimic Source & Human written & 87.66 & 4.30 \\
    \hline
    \end{tabular}
\caption{Performance Comparison of model outputs and human-written sources. SD is the abbreviation ddfor standard deviation.}
\label{tab:code_comparison}
\end{table}

\begin{table}[t]
    \centering
    \setlength{\tabcolsep}{3pt}
    \small
    \begin{tabular}{l|ccccccc}
    \toprule
    Type & pie & line & bar & bar\_num \\
    \midrule
    Percent & 8.0\% & 9.7\% & 8.3\% &3.3\%  \\     
    \toprule
     
    Type  & 3d & area & box  & bubble \\ 
    \midrule
    Percent &5.6\% &3.9\% &4.4\% &2.8\% \\
    \toprule
    Type  &candlestick & funnel & heatmap & multi-axes \\ 
    \midrule
    Percent &2.8\% &2.7\% &3.9\% &3.8\% \\ 
    \toprule
    Type & rader & ring & pie & rose \\ 
    \midrule
    Percent &3.8\% & 2.7\% & 2.8\% & 3.9\% \\ 
    \toprule
    Type & treemap & violin & scatter & quiver \\ 
    \midrule
    Percent & 3.9\% &3.9\% &3.8\% &5.2\%  \\ 
    \toprule
    Type & inset & histogram & graph & error bar \\
    \midrule    
    Percent  &1.2\%  &1.2\% &1.2\% &1.6\% \\ 
    \toprule
    Type & error point & density & Combination & Total \\
    \midrule
    Percent &1.6\% &1.2\%&2.8\% & 100\% \\
    \bottomrule
    \end{tabular}
    \caption{Type distributions of the Chart2Code-160k instruction-tuning dataset.}
    \label{tab:type_analysis}
    \vspace{-10pt}
\end{table}

\subsection{Chart2Code-160k Analysis}
We count the proportion of different charts in the Chart2Code-160k dataset in Table~\ref{tab:type_analysis}. Also, we utilize \texttt{gpt-4o-2024-08-06} to evaluate the quality of the charts in the Chart2Code-160k and compare them with the real-world chart. The prompts are shown in \Cref{fig:chart_quality_prompt}.

\begin{figure*}[t]
    \centering
    \includegraphics[width=\textwidth]{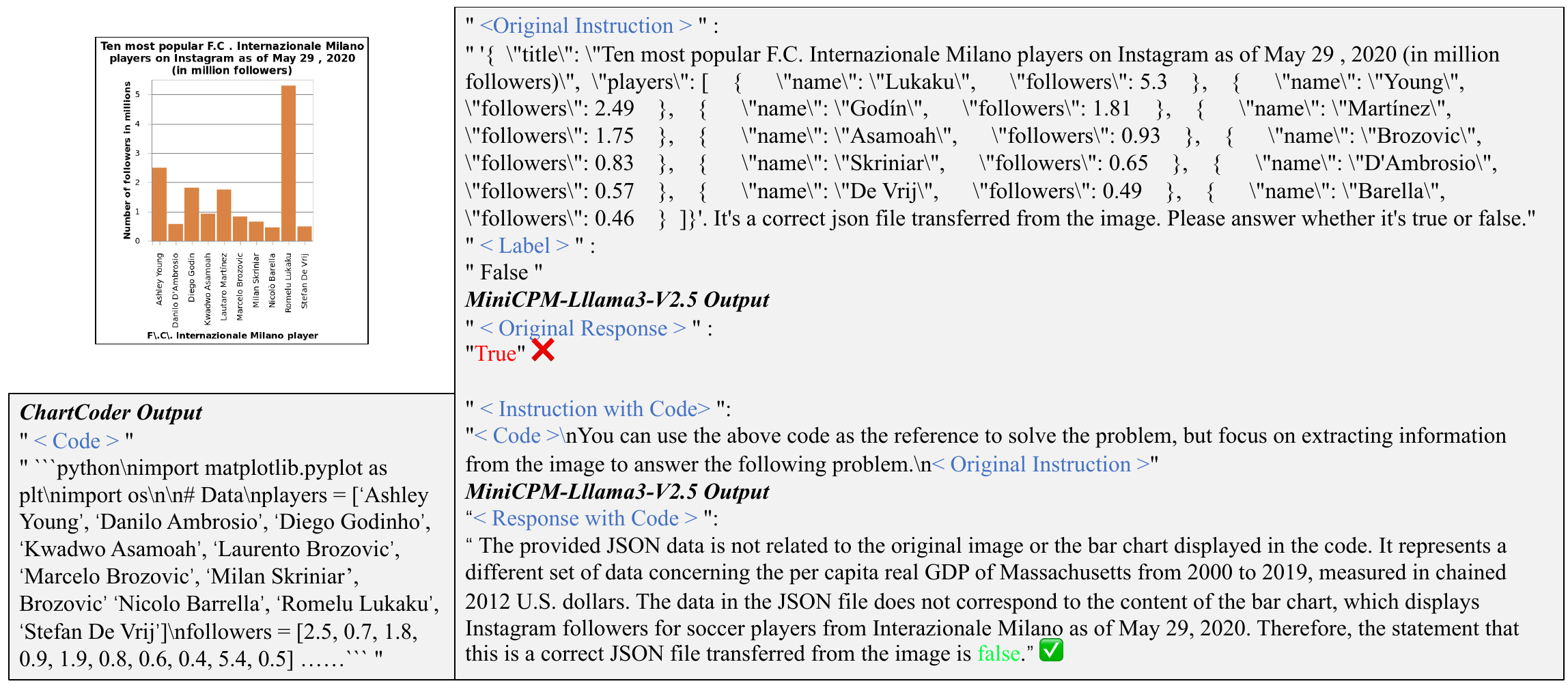}
    \caption{A case study comparing the outputs of utilizing image and image+code as the inputs}
    \label{fig:reasoning}
    \vspace{-10pt}
\end{figure*}

\begin{figure*}[h]
    \centering
    \includegraphics[width=\textwidth]{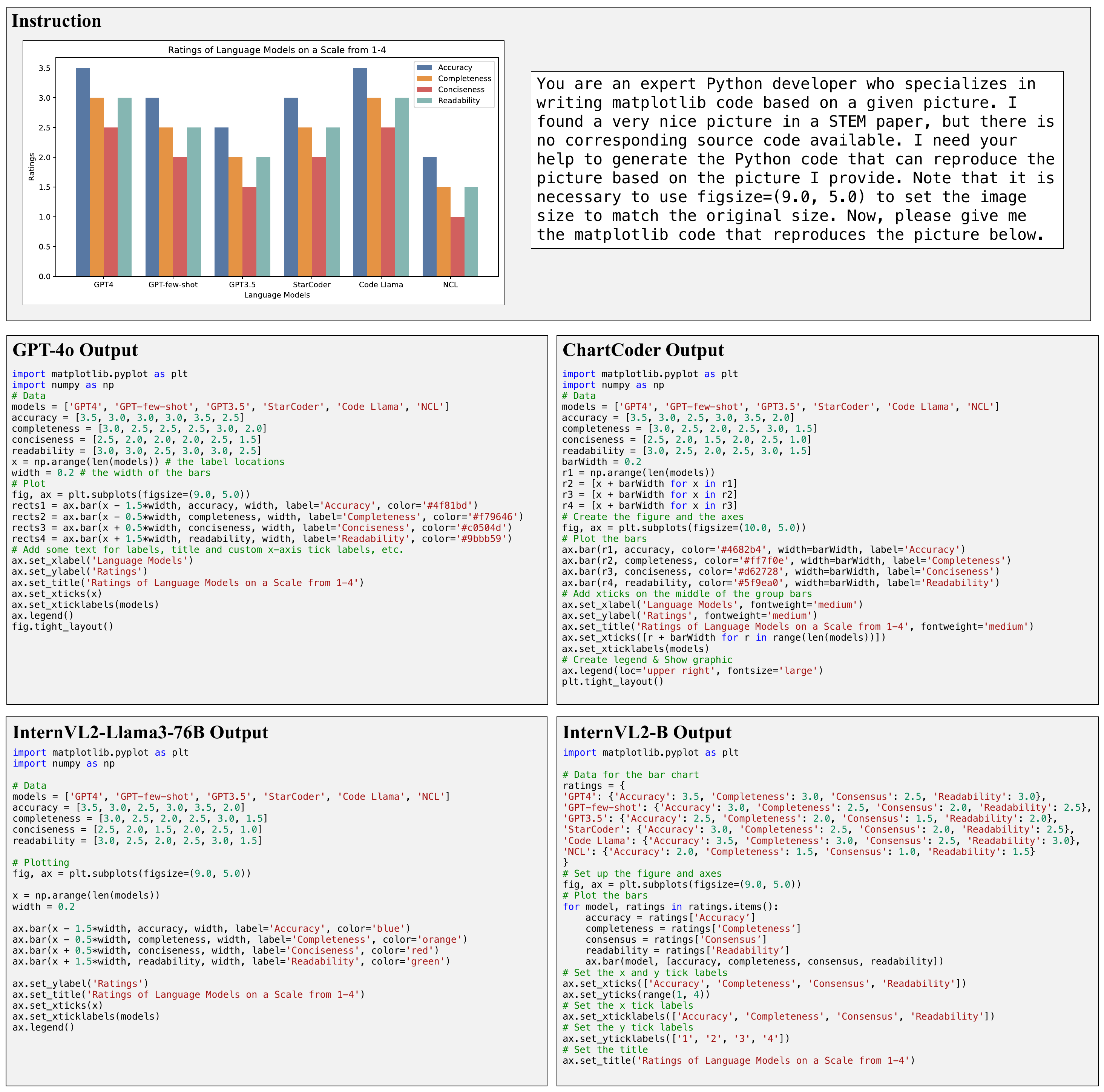}
    \caption{A example of comparing the code corresponding to the bar chart generated by different models.}
    \label{fig:code}
    \vspace{-10pt}
\end{figure*}

\begin{figure*}[t]
\begin{tcolorbox}[colback=white, colframe=black, title=Prompt for Code Readability Evaluation]
Please score the code's readability based on the following four aspects. Each aspect is worth 25 points, for a total of 100 points. \\
Naming Conventions (25 points) \\
Score: [X]/25 \\
Explanation: [Provide a brief explanation of how well the variable, function, and class names convey their purpose and whether the naming style is consistent across the codebase.] \\
Code Structure (25 points) \\
Score: [X]/25 \\
Explanation: [Explain whether functions are concise, whether the code uses indentation and blank lines appropriately, and whether the code is modularized effectively.] \\
Comments (25 points) \\
Score: [X]/25 \\
Explanation: [Discuss the clarity and appropriateness of the comments, and whether functions/methods have proper documentation comments explaining inputs, outputs, and functionality.] \\
Logical Clarity (25 points) \\
Score: [X]/25 \\
Explanation: [Evaluate the intuitiveness of the code, whether it’s easy to understand, and whether the control flow is simple and avoids unnecessary complexity.] \\
Total Score: [X]/100 \\
Summary: [Provide a brief overall assessment of the code’s readability, pointing out strengths and potential areas for improvement.] \\

\end{tcolorbox}
\caption{Prompt for dataset quality evaluation.}
\label{fig:code_read_prompt}
\end{figure*}

\begin{figure*}[t]
\begin{tcolorbox}[colback=white, colframe=black, title=Prompt for Chart Quality Evaluation]
You are a professional chart analyser. Please evaluate the image based on the following four criteria: aesthetics, readability, reproducibility, and data presentation simplicity. Provide a score for each criterion and include an overall score along with a brief evaluation. \\
Scoring Criteria and Requirements:\\
Aesthetics (25 points) \\
Requirements: \\
The chart design should be simple and clear, avoiding complex decorations, and should effectively communicate information. \\
Colors should be harmonious and have high contrast, making it easy to differentiate between different data groups. \\
Legends and labels should be clear, with appropriately sized fonts, avoiding visual clutter. \\
Scoring: [X]/25\\
Readability (30 points)\\
Requirements:\\
The chart should have clear titles, axis labels, and legends, enabling quick communication of the main message.\\
Data curves or point annotations should avoid being overly dense or overlapping, maintaining good readability.\\
The overall layout should follow a logical structure without any confusing elements.\\
Scoring: [X]/30 \\
Reproducibility (30 points)\\
Requirements:\\
The chart design should be easy to replicate using common tools.\\
Data availability is critical: even if the design is simple, missing context or data should result in point deductions.\\
Data should be provided with clear sampling methods, units, and formats, enabling others to recreate the chart from scratch.\\
The presentation of data should align logically with the chart design, avoiding overly customized or complex elements.\\
Scoring: [X]/30 \\
Data Presentation Simplicity (15 points)\\
Requirements:\\
Data presentation should be concise, avoiding redundant information.\\
Data should be displayed in an intuitive way, without excessive curves, points, or annotations.\\
High-scoring charts should focus on the main data being presented, avoiding decorative or unrelated information.\\
Scoring: [X]/15

\end{tcolorbox}
\caption{Prompt for dataset quality evaluation.}
\label{fig:chart_quality_prompt}
\end{figure*}

\end{document}